\documentclass[10pt,twocolumn,letterpaper]{article}

\usepackage{iccv}
\usepackage{times}
\usepackage{epsfig}
\usepackage{graphicx}
\usepackage{amsmath}
\usepackage{amssymb}
\usepackage{algorithm}
\usepackage{algorithmic}
\usepackage{bbm}
\usepackage{amsfonts}
\usepackage{mathrsfs}
\usepackage{multirow}
\usepackage{amsmath,amssymb,amsfonts}
\usepackage{algorithmic}
\usepackage{graphicx}
\usepackage{textcomp}
\usepackage{booktabs}
\usepackage{multicol}
\usepackage{booktabs}
\usepackage{lipsum}
\usepackage{makecell}
\usepackage{caption}
\usepackage{multirow}
\usepackage{placeins}

\usepackage[pagebackref=true,breaklinks=true,letterpaper=true,colorlinks,bookmarks=false]{hyperref}

\iccvfinalcopy 

\def\iccvPaperID{406} 

\begin{document}

\title{Med3DVLM: An Efficient Vision-Language Model for 3D Medical Image Analysis}

\author{Yu Xin{$^{1\dag}$},~ Gorkem Can Ates{$^{1\dag}$},~ Kuang Gong{$^{1}$}, ~ Wei Shao{$^{1*}$}\\\vspace{-8pt}{\small~}\\
{$^{1}$}University of Florida\\
{\small{{$^{\dag}$}Contribute equally~~{$^{*}$}Corresponding to: \tt{weishao@ufl.edu}}}
}

\maketitle
\footnotetext[1]{This work has been submitted to the IEEE for possible publication. Copyright may be transferred without notice, after which this version may no longer be accessible.}
\begin{abstract}
Vision-language models (VLMs) have shown promise in 2D medical image analysis, but extending them to 3D remains challenging due to the high computational demands of volumetric data and the difficulty of aligning 3D spatial features with clinical text. We present Med3DVLM, a 3D VLM designed to address these challenges through three key innovations: (1) DCFormer, an efficient encoder that uses decomposed 3D convolutions to capture fine-grained spatial features at scale; (2) SigLIP, a contrastive learning strategy with pairwise sigmoid loss that improves image-text alignment without relying on large negative batches; and (3) a dual-stream MLP-Mixer projector that fuses low- and high-level image features with text embeddings for richer multi-modal representations. We evaluated our model on the M3D dataset, which includes radiology reports and VQA data for 120,084 3D medical images. The results show that Med3DVLM achieves superior performance on multiple benchmarks. For image-text retrieval, it reaches 61.00\% R@1 on 2,000 samples, significantly outperforming the current state-of-the-art M3D-LaMed model (19.10\%). For report generation, it achieves a METEOR score of 36.42\% (vs. 14.38\%). In open-ended visual question answering (VQA), it scores 36.76\% METEOR (vs. 33.58\%), and in closed-ended VQA, it achieves 79.95\% accuracy (vs. 75.78\%). These results demonstrate Med3DVLM’s ability to bridge the gap between 3D imaging and language, enabling scalable, multi-task reasoning across clinical applications. Our code is publicly available at \url{https://github.com/mirthAI/Med3DVLM}.
\end{abstract}

\section{Introduction}
\label{sec:introduction}
Medical image analysis plays a crucial role in diagnosing and treating diseases such as cancer, cardiovascular conditions, and neurological disorders. However, existing models are often task-specific, lack adaptability to new tasks, and do not support real-time user interactions.  Vision-language models (VLMs), such as Contrastive Language-Image Pre-Training (CLIP)~\cite{radford2021learning} and Large Language and Vision Assistant (LLaVA)~\cite{liu2023visual}, offer greater versatility in dynamic clinical settings by aligning medical images with textual reports. CLIP leverages contrastive learning on image-text pairs to enable zero-shot image classification and image-text retrieval. LLaVA extends CLIP by integrating its visual encoder with a large language model (LLM), facilitating interactive tasks such as report generation and visual question answering (VQA). Despite their success in analyzing 2D medical images, such as chest X-rays~\cite{endo2021retrieval} and pathology slides~\cite{lu2024multimodal}, their application to 3D imaging remains limited.

3D imaging modalities, such as computed tomography (CT) and magnetic resonance imaging (MRI), provide volumetric data that capture spatial details unavailable in 2D images. However, extending VLMs to 3D presents significant challenges. First, a 3D scan consists of hundreds of slices, making slice-by-slice analysis prone to losing global context. Second, directly building 3D VLMs significantly increases computational complexity due to the higher dimensionality. Third, the scarcity of publicly available 3D image-report pairs further limits model development. 

Several VLMs have attempted to bridge this gap, but each has limitations in the 3D setting. For example, PMC-CLIP~\cite{lin2023pmc}, trained on large-scale biomedical literature images, is restricted to 2D inputs, leading to poor performance on 3D image understanding. RadFM~\cite{wu2023towards} unifies 2D and 3D data but is primarily optimized for text generation tasks, such as VQA, and struggles with broader image-text understanding. More recently, M3D-LaMed~\cite{bai2024m3d} was introduced as a generalist multi-modal model for 3D medical image analysis. By combining a CLIP-pretrained 3D visual encoder with a 3D spatial pooling perceiver, M3D-LaMed enables direct reasoning on 3D scans and achieves state-of-the-art performance across multiple benchmarks, including image-text retrieval, report generation, open-ended and closed-ended VQA, as well as 3D segmentation and localization tasks.

Despite these advancements, M3D-LaMed has notable limitations. Its 3D backbone, similar to many other 3D VLMs, incurs high computational costs when processing high-resolution 3D volumes, as standard 3D vision transformers scale poorly with image size. Additionally, M3D-LaMed's vision-language alignment relies on CLIP's contrastive loss, which compares each image-text pair against a large set of negative pairs in the batch. This approach works well for large datasets but can be less effective in smaller medical datasets, where meaningful negative samples are limited, and similar images may still share semantic information. Furthermore, its multi-modal fusion mechanism, which projects visual features into the LLM via a multilayer perceptron (MLP), may not sufficiently capture complex cross-modal interactions.

To overcome the limitations of existing VLMs in 3D imaging, we introduce Med3DVLM, a novel VLM that incorporates three key innovations to improve 3D feature learning, cross-modal alignment, and multi-modal projection:
\begin{enumerate}
    \item \textbf{Efficient 3D Feature Encoding.} We integrate DCFormer~\cite{ates2025dcformer}, a 3D image encoder that efficiently captures volumetric features by decomposing 3D convolutions into three parallel 1D convolutions along the depth, height, and width axes. This approach reduces computational complexity and enables richer, more scalable feature representations for 3D image volumes.
    \item \textbf{Improved Vision-Language Alignment.} We adopt SigLIP~\cite{zhai2023sigmoid}, a sigmoid-based language-image pretraining scheme, to improve image-text alignment. Unlike CLIP's softmax-based contrastive loss, which relies on distinguishing positive pairs from a large batch of negative samples, SigLIP uses a pairwise sigmoid loss that directly optimizes each image-text pair independently. This eliminates the need for global similarity normalization across batches, making training more stable and less sensitive to batch size. 
    \item \textbf{Multi-Scale Multi-Modal Projector.} We introduce a novel projector based on MLP-Mixer~\cite{tolstikhin2021mlp} to fuse image and text embeddings effectively. Utilizing a low-high-hybrid design, it blends detailed low-level and abstract high-level features from the image encoder with LLM text embeddings using stacked MLP layers. Inspired by MLP-Mixer’s ability to mix spatial and feature data, this dual-stream approach captures richer cross-modal interactions than simple linear projection, improving the LLM’s decoding accuracy.
\end{enumerate}

\section{Related Work}
\subsection{Medical Vision-Language Models}

Medical VLMs utilize multi-modal learning to enhance tasks such as disease diagnosis, report generation, and medical VQA through improved integration and understanding of medical images and text.
One of the earliest efforts, ConVIRT~\cite{zhang2022contrastive}, applied contrastive learning to learn visual representations from paried image-text data, outperforming ImageNet pretraining in medical image classification and zero-shot retrieval tasks. Similarly, BioViL~\cite{boecking2022making} leveraged large-scale biomedical datasets to refine multi-modal representations, significantly outperforming previous supervised methods. Further extending these capabilities, MedCLIP~\cite{wang2022medclip} replaced the InfoNCE loss with semantic matching loss based on medical knowledge, demonstrating its ability to learn generalized representations with limited data. 

Recent efforts have aimed to develop VLMs for 3D medical imaging.
RadFM~\cite{wu2023towards} introduced a generalist VLM trained on massive datasets that consisted of both 2D and 3D medical images and associated radiology reports. It integrated contrastive learning and generative modeling across diverse imaging modalities (X-ray, CT, MRI) and textual reports, enabling a unified radiology representation. Extending this, M3D~\cite{bai2024m3d} introduced a large-scale dataset, M3D-Data, along with M3D-LaMed, a multi-modal vision-language model for 3D medical image analysis tasks such as image-text retrieval, report generation, VQA, and promotable segmentation. Its successor, E3D-GPT~\cite{lai2024e3d}, employed a 3D multi-modal masked autoencoder framework to further enhance image-text retrieval, leading to improved performance in report generation and VQA. In parallel, researchers have also explored text-guided 3D medical image segmentation~\cite{zhao2023one,huang2024cat,xin2025text3dsam}, where language prompts are used to localize anatomical structures or pathologies. 

\subsection{Radiology Report Generation}
Automated report generation aims to produce descriptive, accurate, and clinically relevant reports from medical images. It improves diagnostic accuracy while reducing radiologists' workload and supporting care in resource-limited settings. Early approaches primarily adopted an encoder-decoder architecture, where convolutional neural networks (CNNs) extracted image features, and long short-term memory (LSTM) networks generated text descriptions~\cite{xue2018multimodal,harzig2019addressing,yuan2019automatic,wang2021self}. However, these methods struggled with long-range dependencies, often produced repetitive text, and had limited capacity to capture complex medical semantics.

The transformer architecture~\cite{vaswani2017attention} addressed key limitations of CNNs and LSTMs by using self-attention to capture global dependencies. Pretrained LLMs~\cite{touvron2023llama,penedo2023refinedweb} based on transformers are now adapted into medical VLMs for report generation.  R2GenGPT~\cite{wang2023r2gengpt} introduced an efficient visual alignment module to better integrate image features with LLM word embeddings, improving text coherence and clinical relevance. Med-Flamingo~\cite{moor2023med}, based on OpenFlamingo-9B, was pretrained on interleaved medical image-report pairs, achieving superior performance in generating clinically useful responses. More recent methods, such as LLaVA-Med, were based on the LLaVA~\cite{liu2023visual} framework, which combined a visual encoder with an LLM to generate detailed and coherent reports. CT2Rep~\cite{hamamci2024ct2rep} proposed an advanced 3D vision encoder to generate radiology reports for 3D medical imaging, specifically targeting chest CT volumes. Similarly, CT-CHAT~\cite{hamamci2024developing} adapted the LLaVA framework for chest CT report generation, demonstrating the effectiveness of large-scale pretrained LLMs in 3D medical imaging.

\subsection{Medical Visual Question Answering}
A medical VQA system can answer natural language questions about medical images. This task can be closed-ended (with answers like yes/no or a choice from a fixed list) or open-ended (free-form text answers). Early work on medical VQA largely treated it as a classification problem, where models selected the correct answer from a predefined set of possible responses~\cite{nguyen2019overcoming,zheng2020learning,gupta2021hierarchical,liu2021contrastive}. These models typically encoded images (e.g., using CNNs) and text (e.g., using transformers), and then mapped the combined features to a predefined answer space. Although this approach worked well for simple questions, such as identifying an organ, it struggled with questions requiring detailed explanation or answers not included in the predefined list. 

Nowadays, approaches to medical VQA increasingly adopt generative architectures, treating answer generation as a sequence prediction task, which enables open-ended responses. An early step in this direction was CGMVQA~\cite{ren2020cgmvqa}, which added a generative decoder branch alongside the traditional classifier. Recent VQA systems have fully transitioned to generative models~\cite{dong2025generative}. 
These models (e.g., Med-PaLM M~\cite{tu2024towards} and LLaVA-Med~\cite{li2023llava}) extend an LLM with visual inputs via visual instruction tuning~\cite{liu2023visual}, allowing the LLM to interpret medical images and answer questions in a conversational manner. 
Recently, a few works have explored VQA on 3D medical images~\cite{bai2024m3d,hamamci2024developing}, where models can directly perceive entire 3D image volumes and generate context-aware responses, offering more accurate and holistic medical insights.

\section{Methods}

\subsection{Dataset}
This study used the publicly available M3D dataset~\cite{bai2024m3d}, collected with informed consent and ethical approval by the original investigators. All data were de-identified, and no additional consent or approval was required for this secondary analysis.
We used two subsets of the M3D dataset: M3D-Cap and M3D-VQA. M3D-Cap includes 120K image-text pairs, while M3D-VQA contains 662K instruction-response pairs. The dataset can support tasks such as image-text retrieval, radiology report generation, and VQA. All image volumes were resampled to a fixed size of 128x256×256. The dataset was divided into training (115k samples), validation (3k samples), and test (2k samples) sets.

\subsection{Med3DVLM}
\begin{figure*}[!ht]
  \centering
  \includegraphics[width=\textwidth]{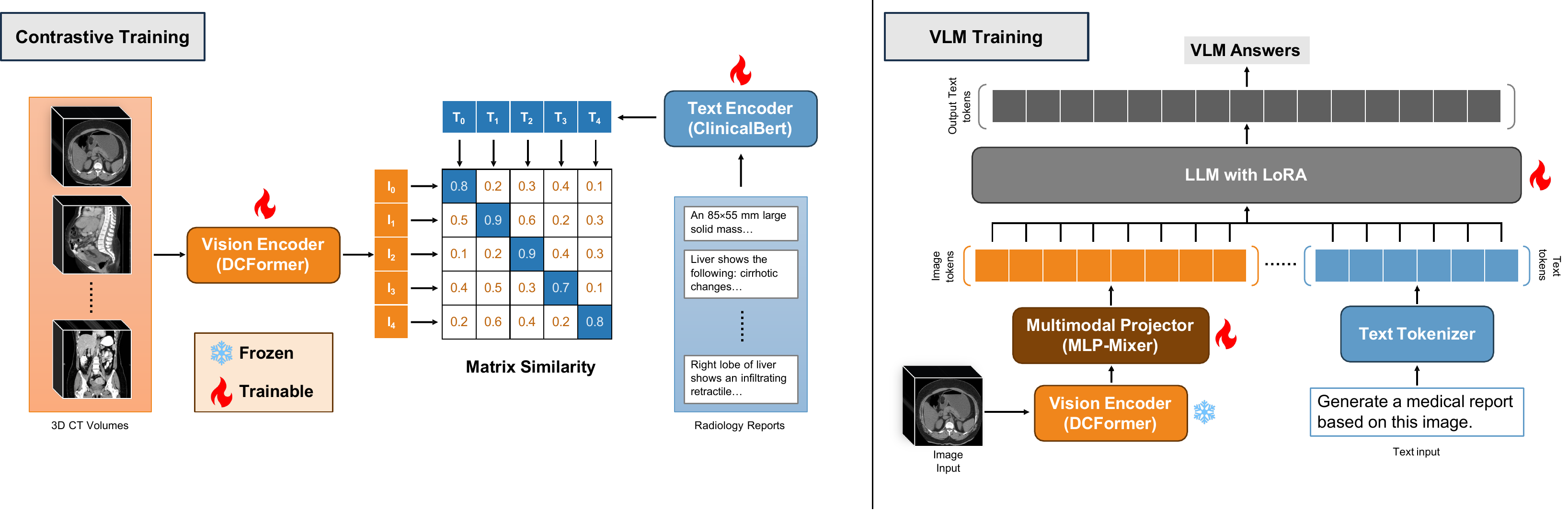}
  \caption{Overview of our Med3DVLM model.}
  \label{fig:pipeline}
\end{figure*}

We introduce Med3DVLM, a 3D medical VLM consisting of three core components: a vision encoder, a multi-modal projector, and a large language model. The vision encoder extracts detailed visual features from 3D medical image volumes. These features are integrated with text embeddings through the multi-modal projector, facilitating cross-modal interactions. The LLM then generates coherent, contextually accurate outputs based on these fused features. The overall model architecture is illustrated in Figure \ref{fig:pipeline}. The training process is divided into three stages: (1) contrastive pretraining, (2) multi-modal projector pretraining, and (3) VLM fine-tuning.

\subsubsection{Contrastive Pretraining}
Following the CLIP framework~\cite{radford2021learning}, the vision and text encoders are trained jointly using a contrastive loss to align 3D images with their corresponding radiology reports. The goal is to maximize the similarity between embeddings of positive (matching) image-text pairs while minimizing it for negative (non-matching) pairs.

For the vision encoder, we utilize DCFormer~\cite{ates2025dcformer}, which efficiently extracts volumetric features using decomposed 3D convolutions. 
This approach mitigates computational challenges in high-resolution 3D image analysis, where traditional 3D CNNs and vision transformers struggle due to the cubic scaling of 3D CNNs with kernel size and the quadratic scaling of vision transformers with image size. DCFormer decomposes 3D convolutions into three 1D parallel depthwise convolutions:
\begin{align}
X^*_{\text{h}} &= \text{DWConv}^{k_h \times 1 \times 1}_{C \rightarrow C} (X), \\
X^*_{\text{w}} &= \text{DWConv}^{1 \times k_w \times 1}_{C \rightarrow C} (X), \\
X^*_{\text{d}} &= \text{DWConv}^{1 \times 1 \times k_d}_{C \rightarrow C} (X), 
\end{align}
where $X \in \mathbb{R}^{B \times C \times H \times W \times D}$ denote the input feature map, \( B \) is the batch size, \( C \) is the number of channels, and \( H \), \( W \), and \( D \) represent the spatial dimensions (height, width, and depth), respectively. The kernel sizes in these dimensions, denoted as \( (k_h, k_w, k_d) \), are set by default to $k_h = k_w = k_d = k \in \{13, 11, 9, 7\}$ to leverage large kernels for improved feature extraction. The extracted features are then normalized and summed to produce the final output:
\begin{align}
X' &= X + \text{Norm}_h(X^*_{\text{h}}) + \text{Norm}_w(X^*_{\text{w}}) + \text{Norm}_d(X^*_{\text{d}}).
\end{align}
Such a decomposition strategy enables the processing of 3D volumes at a larger size of 128x256×256 in this study, preserving fine-grained spatial details for improved image-text alignment. DCFormer, short for DeComposed Former, derives its name from this decomposition strategy. The “Former” suffix is inspired by the MetaFormer~\cite{yu2022metaformer} framework, which generalizes Transformer-like architectures by replacing attention mechanisms with alternative token-mixing operations.

In this work, we utilize DCFormer-small variant, which consists of a stem stage and four hiararchical stages with [2, 3, 6, 2] layers and [96, 192, 384, 768] channels~\cite{ates2025dcformer}. This adaptation of DCFormer within the LLaVA~\cite{liu2023visual} framework enables efficient processing of high-resolution 3D volumes while preserving spatial detail. By reducing computational complexity, it supports scalable 3D medical vision-language modeling under typical GPU constraints and addresses key limitations of 3D vision transformers and CNNs.

For the text encoder, we use ClinicalBERT~\cite{wang2023optimized}, an adaptation of BERT~\cite{devlin2019bert} pretrained on clinical notes. Designed for clinical text, ClinicalBERT is well-suited for processing radiology reports, generating embeddings that effectively capture their semantic content.

Traditional softmax-based contrastive learning (e.g., CLIP) distinguishes positive pairs from a large batch of negative samples. While effective for large datasets, this approach struggles with smaller medical datasets, where meaningful negative samples are limited, and similar images may share semantic information. To address this, we adopt SigLIP~\cite{zhai2023sigmoid}, a sigmoid-based language-image pretraining scheme that directly optimizes each image-text pair independently. This eliminates the need for global similarity normalization across batches, resulting in more stable training and reduced sensitivity to batch size. Sigmoid loss is defined as:

\begin{equation}
  \mathcal{L}_{\text{SigLIP}} = - \frac{1}{|\mathcal{B}|} \sum_{i=1}^{|\mathcal{B}|} \sum_{j=1}^{|\mathcal{B}|} \log \underbrace{\frac{1}{1 + e^{z_{ij}(-t \mathbf{x}_i \cdot \mathbf{y}_j + b)}}}_{\mathcal{L}_{ij}},
\end{equation}
where \(z_{ij}\) is the label for the \(i\)-th image and \(j\)-th text pair, \(z_{ij} = 1\) for positive pairs and \(z_{ij} = -1\) otherwise. The sigmoid loss \(\mathcal{L}_{ij}\) is computed for each image-text pair, with \(t\) as the temperature parameter, \(\mathbf{x}_i\) and \(\mathbf{y}_j\) as the image and text embeddings, respectively, and \(b\) as a bias term. The overall loss is averaged over all pairs in the batch. 

For comparison, the CLIP loss is given by:
\begin{equation}
\mathcal{L}_{\text{CLIP}} = -\frac{1}{2|\mathcal{B}|} \sum_{i=1}^{|\mathcal{B}|}
\left(
\underbrace{\log \frac{e^{t \mathbf{x}_i \cdot \mathbf{y}_i}}{\sum_{j=1}^{|\mathcal{B}|} e^{t \mathbf{x}_i \cdot \mathbf{y}_j}}}_{\text{image} \rightarrow \text{text softmax}}
+
\underbrace{\log \frac{e^{t \mathbf{x}_i \cdot \mathbf{y}_i}}{\sum_{j=1}^{|\mathcal{B}|} e^{t \mathbf{x}_j \cdot \mathbf{y}_i}}}_{\text{text} \rightarrow \text{image softmax}}
\right)
\end{equation}

Unlike traditional medical VLMs that rely on CLIP pretraining and require large batch sizes for effective contrastive learning, SigLIP adopts a pairwise sigmoid loss that operates independently on each image-text pair. Rather than comparing one positive against many negatives using softmax, SigLIP formulates contrastive learning as a binary classification problem. Matched pairs are pushed together, and unmatched pairs are pushed apart. This design removes the need for batch-wide normalization, improves alignment stability, and reduces sensitivity to batch size, which makes it more suitable for small-batch, semantically rich medical imaging settings.

\subsubsection{Multi-modal Projector Pretraining}
We propose an effective multi-modal projector for the interaction between visual and textual embeddings. In this pretraining stage, all model weights are frozen except for those in the projector. To effectively capture spatial and feature dependencies, we adopt the MLP-Mixer architecture~\cite{tolstikhin2021mlp}, which alternates between a token-mixing MLP (spatial mixer) and a channel-mixing MLP. 
 
Given an input image feature map \(X \in \mathbb{R}^{n \times d}\), where \(n\) is the number of tokens and \(d\) is the embedding dimension, the MLP-Mixer operations are defined as:
\begin{align}  
  U &= W_2\,\sigma(W_1\,(\text{Norm}(X)^\top)),\\
  Y &= W_4\,\sigma(W_3\,\text{Norm}(U^\top)),
\end{align}
where \( W_1, W_2, W_3, W_4 \) are learnable weight matrices, and \( \sigma \) denotes a nonlinear activation function (e.g., GELU). The token-mixing MLP captures interactions across spatial tokens, while the channel-mixing MLP models dependencies across feature channels.

\begin{figure}[!ht]
  \centering
  \includegraphics[width=0.5\textwidth]{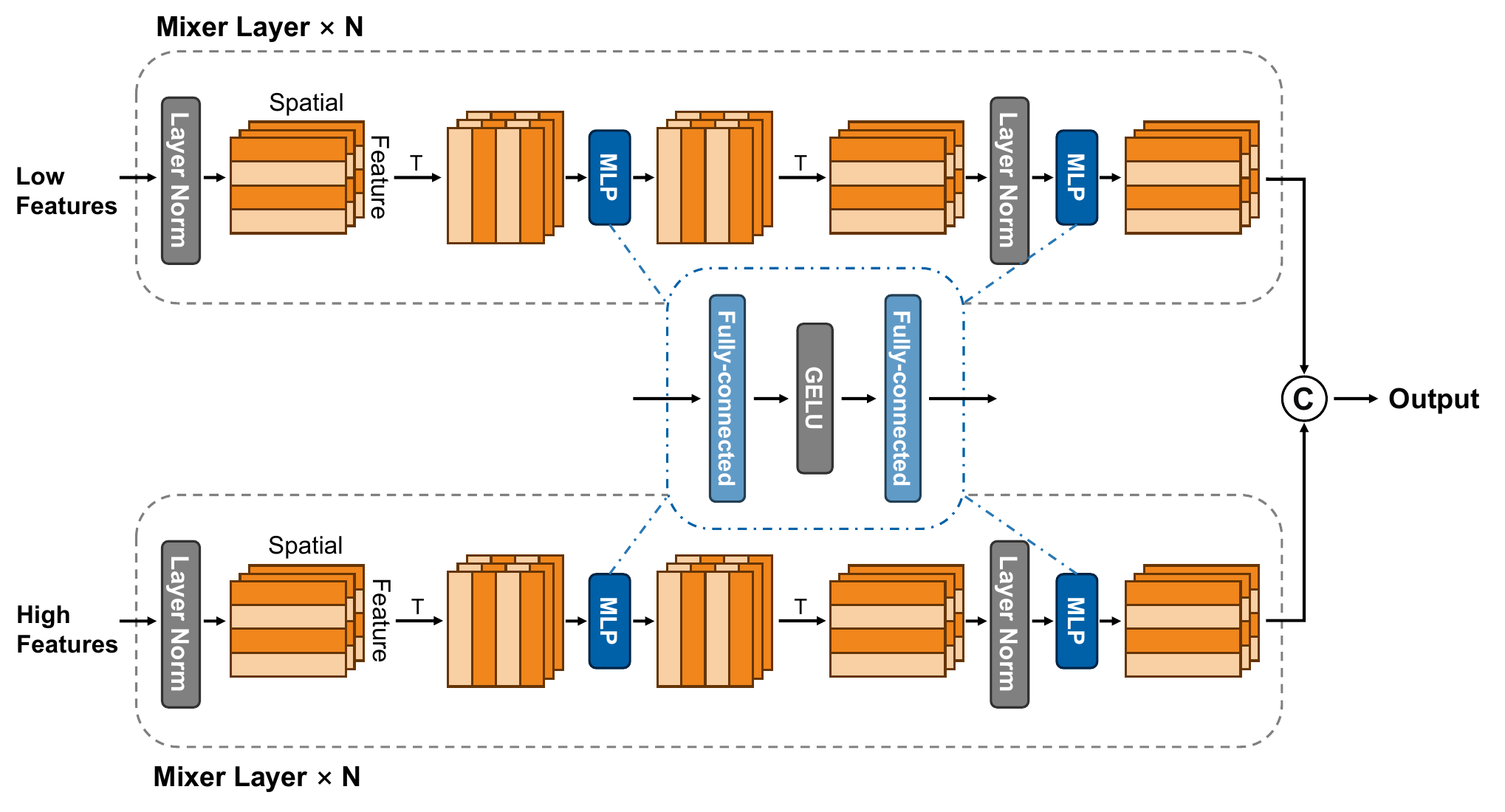}
  \caption{Multi-modal projector based on multi-scale MLP-Mixer.}
  \label{fig:mm_projector}
\end{figure}

To further enhance multi-modal fusion, we propose a low-high hybrid Mixer-MLP architecture (Figure~\ref{fig:mm_projector}), inspired by Janus Pro~\cite{chen2025janus}. Unlike M3D-LaMed~\cite{bai2024m3d}, which produces large spatial outputs \((B, 2048, 768)\) that require lossy downsampling, DCFormer outputs semantically rich features \((B, 32, 768)\) from its final layer and spatially detailed features \((B, 256, 384)\) from its penultimate layer. Because of the compact spatial dimensions of DCFormer’s outputs, our model can adopt the low-high hybrid structure without exceeding the token length limit of the LLM. This is crucial, as excessively long image token sequences can crowd out text tokens, leading to truncated questions or reports during inference.
To leverage both types of features, we utilize two parallel MLP-Mixer modules—one for each layer. After passing through \(N\) Mixer layers, the resulting feature outputs are concatenated into a unified sequence of image tokens. In parallel, the input text is tokenized into a sequence of token IDs using the vocabulary of the LLM. This step follows the LLaVA~\cite{liu2023visual} framework, which requires token embeddings for multimodal fusion. The resulting image and text tokens are then fused and passed to the LLM for joint reasoning.

This hybrid design enables the projector to fuse high-level abstract semantics with low-level spatial details, improving multi-modal alignment. Unlike prior approaches that downsample features at the expense of context, our method preserves diverse information without increasing token length. As a result, our projector produces richer and more semantically aligned joint representations, significantly enhancing radiology report generation and VQA. To the best of our knowledge, this work is the first to introduce a dual stream low-high hybrid MLP-Mixer for multimodal fusion in a 3D vision-language model. Unlike the original single-stream MLP-Mixer, our design processes both low-level spatial features and high-level semantic representations extracted from the 3D image encoder. This configuration enables the model to align textual inputs with both fine-grained anatomical details and high-level semantic context, which is critical for medical vision-language understanding. Empirically, it outperforms standard single-scale MLP-Mixer across multiple tasks.

We conducted this pretraining stage on both the M3D-Cap and M3D-VQA datasets. To mitigate binary-response bias and ensure richer semantic grounding, we excluded yes/no questions from M3D-VQA.

\subsubsection{VLM Fine-tuning}
In this stage, we fine-tune the multi-modal projector and the LLM for the report generation and VQA. Instead of fine-tuning the entire LLM, we adopt Low-Rank Adaptation (LoRA)~\cite{hu2022lora}, which significantly reduces the number of trainable parameters by introducing low-rank updates to the model’s weights. This approach enables efficient adaptation while preserving the LLM's general knowledge.

Specifically, LoRA modifies a weight matrix \( W_0 \in \mathbb{R}^{d \times k} \) by adding a learnable low-rank update:
\begin{equation}
W = W_0 + \Delta W, \quad \text{where} \quad \Delta W = AB
\end{equation}
where \( A \in \mathbb{R}^{d \times r} \) and \( B \in \mathbb{R}^{r \times k} \) are low-rank matrices with \( r \ll \min(d, k) \). This decomposition allows task-specific tuning using far fewer parameters than full fine-tuning, resulting in faster training and lower memory usage.

During fine-tuning, only the weights of the multi-modal projector and the LoRA modules are updated, while all other parameters of the LLM remain frozen. The LLM is initialized with Qwen2.5-7B-Instruct~\cite{qwen2.5}, a model pretrained for instruction-following tasks. By jointly optimizing the LoRA parameters and the multi-modal projector, our model effectively adapts to the characteristics of the M3D-Cap and M3D-VQA datasets.

\subsection{Evaluation Metrics}

We evaluate image-text retrieval performance using Recall@K (R@K) metrics, which quantify the percentage of correct matches within the top K retrieved items. Specifically, R@1 represents the proportion of queries where the correct match is ranked first, while R@5 and R@10 indicate the proportion of queries where the correct match appears within the top 5 or 10 results, respectively.

We evaluate the quality of radiology report generation and open-ended VQA using four key metrics: BLEU~\cite{papineni2002bleu}, ROUGE~\cite{lin2004rouge}, METEOR~\cite{banerjee2005meteor}, and BERTScore~\cite{BERT-Score}. BLEU measures precision by evaluating the overlap of n-grams between the generated text and the reference text, focusing on how many words or phrases in the generated report are found in the reference text. The BLEU score is calculated as:
\begin{equation}
  \text{BLEU} = BP \cdot \exp\left(\sum_{n=1}^{N} w_n \cdot \log p_n\right)
\end{equation}
where BP (brevity penalty) is defined as:
\begin{equation}
  \text{BP} = \begin{cases}
    1 & \text{if } c > r \\
    e^{(1 - \frac{r}{c})} & \text{if } c \leq r
  \end{cases}
\end{equation}
Here, \(c\) is the length of the generated text, \(r\) is the length of the reference text, \(w_n\) is the weight for n-grams, and \(p_n\) is the precision for n-grams of size \(n\).

ROUGE emphasizes recall, assessing how much of the reference text appears in the generated output, which is particularly useful for summarization tasks to ensure key details are retained. The ROUGE score is calculated as:
\begin{equation}
  \text{ROUGE} = \frac{\sum_{n=1}^{N} \text{Recall}_n}{\sum_{n=1}^{N} \text{Reference}_n}
\end{equation}
where \(\text{Recall}_n\) is the recall of n-grams in the generated report, and \(\text{Reference}_n\) is the reference n-grams.

METEOR improves upon BLEU and ROUGE by considering synonyms and stemming, balancing precision and recall to provide a more flexible and comprehensive measure. The METEOR score is calculated as:
\begin{equation}
  \text{METEOR} = \frac{\text{Precision} \cdot \text{Recall}}{\text{Precision} + \text{Recall}}
\end{equation}
where Precision and Recall are calculated based on the number of matching words between the generated and reference reports.

BERTScore utilizes deep learning-based contextual embeddings to evaluate semantic similarity. The BERTScore is calculated as:
\begin{equation}
  \text{BERT-Score} = \frac{1}{N} \sum_{i=1}^{N} \frac{\sum_{j=1}^{M} \text{cosine}(e_i, e_j)}{M}
\end{equation}
where \(e_i\) and \(e_j\) are the contextual embeddings of the generated and reference words, respectively, and \(M\) is the number of reference words.

\subsection{Implementation Details}  

We implemented Med3DVLM using PyTorch~\cite{paszke2019pytorch} and Hugging Face Transformers~\cite{wolf2019huggingface}, leveraging DeepSpeed ZeRO2 and BF16 precision on 8 NVIDIA A100 80GB GPUs. We used the AdamW optimizer~\cite{loshchilov2017decoupled} with a warmup ratio of 0.03 and a cosine learning rate scheduler.

\subsubsection{Contrastive Pretraining} During contrastive pretraining on the M3D-Cap dataset, we used image-text pairs, where images were 3D CT volumes and texts were corresponding radiology reports. We computed similarity scores between pooled image and text embeddings to optimize a contrastive loss. We resized 3D CT volumes to \(128 \times 256 \times 256\). We used DCFormer-small~\cite{ates2025dcformer} as the vision encoder, and ClinicalBERT~\cite{wang2023optimized} as the text encoder. Image features were aggregated via mean pooling, while text features were achieved from the [CLS] token, as it captures the global semantic representation of the entire sequence in ClinicalBERT, making it suitable for contrastive alignment with image features. Both were projected to a shared 768-dimensional space. We used a batch size of 64, learning rate of \(1 \times 10^{-4}\), weight decay of 0.1, and trained for 100 epochs.

\subsubsection{VLM Pretraining} For VLM pretraining, we used image-question-answer triplets, where the image was 3D CT volume, the question was a natural language question query, and the answer was the corresponding free-text response. These triplets were sourced from the M3D-Cap and M3D-VQA datasets (excluding yes/no questions) dataset. The DCFormer-small encoder was kept frozen. To align the image encoder and the LLM, we introduced a hybrid MLP-Mixer projector that fused low-level features from the penultimate encoder layer (\(256 \times 384\)) and high-level features from the final layer (\(32 \times 768\)) of the image encoder. Each feature stream passed through \(N\) Mixer layers, and the resulting tokens were concatenated and paired with text tokens as input to the LLM. This stage used a batch size of 16, learning rate of \(1 \times 10^{-4}\), no weight decay, and was trained for 3 epochs.

\subsubsection{VLM Fine-tuning}
For VLM fine-tuning, We continued training using image–question–answer triplets from M3D-Cap and M3D-VQA, now including yes/no questions. The image encoder and MLP-Mixer projector were reused. The model was fine-tuned to generate answers conditioned on the image and question. We used LoRA~\cite{hu2022lora} with a rank of 16, scaling factor \(\alpha = 32\), and dropout of 0.05, updating only the LoRA modules and the projector while keeping all other weights frozen. We fine-tuned the model using a batch size of 8, learning rate of \(5 \times 10^{-5}\), no weight decay, and training for 5 epochs. All M3D-LaMed results are reported as in the original paper~\cite{bai2024m3d}.

\subsubsection{Inference and Evaluation}
During inference, we evaluated two tasks: report generation and visual question answering. In both tasks, the input consisted of a 3D CT volume and a natural language prompt. For report generation, the prompt was a fixed instruction such as “Please provide a radiology report for the given CT volume.” The model generated a free-text radiology report (see Fig \ref{fig:report_examples} for an example). In VQA, the prompt was a question about the image. In open-ended VQA, the model generated free-form answers. In close-ended VQA, multiple answer options were included in the prompt, and the model selected the most appropriate ones (see Fig \ref{fig:vqa_examples} for examples).

\section{Results}

\subsection{Image-Text Retrieval}

Med3DVLM significantly outperforms M3D-LaMed in image-to-text and text-to-image retrieval tasks, as shown in Table~\ref{tab:retrieval_main}. It achieves the highest Recall@1 across all test set sizes (100, 500, 1000, and 2000). These consistent gains highlight its ability to learn semantically rich representations from high-resolution volumetric data and align them with corresponding text descriptions.

Notably, the performance gap between Med3DVLM and M3D-LaMed widens with larger test sets, suggesting that Med3DVLM generalizes better under more challenging and diverse retrieval scenarios. This trend indicates robustness not only to variation in anatomical content and imaging conditions but also to increasing data scale—a critical requirement for deployment in real-world clinical settings. In such environments, retrieval systems must support large-scale databases while maintaining high precision.

\begin{table*}[!ht]
  \caption{Image-text retrieval performance on the M3D dataset. R@K: Recall@Top K. IR: image retrieval. TR: text retrieval. DCFormer-S: small version of DCFormer.}
    \centering
    \small
    \begin{tabular}{c|c|rrrr|rrrr|rrrr}
      \toprule 
        \multicolumn{2}{c|}{Methods} & \multicolumn{4}{c|}{PMC-CLIP} & \multicolumn{4}{c|}{M3D-LaMed} & \multicolumn{4}{c}{DCFormer-S SigLIP (Ours)} \\
      \midrule
        \multicolumn{2}{c|}{Test Samples} & 100 & 500 & 1000 & 2000 & 100 & 500 & 1000 & 2000 & 100 & 500 & 1000 & 2000 \\
      \midrule
        \multirow{3}{*}{IR} & R@1 & 9.00  & 4.40  & 1.90  & 1.15  & 64.00  & 39.60  & 27.30  & 19.10  & \textbf{92.00}  & \textbf{77.20}  & \textbf{69.30}  & \textbf{61.00}  \\
        & R@5 & 28.00  & 12.80  & 7.60  & 4.35  & \textbf{95.00}  & 76.20  & 61.10  & 47.45  & \textbf{95.00}  & \textbf{92.80}  & \textbf{91.40}  & \textbf{87.20}  \\ 
        & R@10 & 45.00  & 18.80  & 12.10  & 7.60  & \textbf{99.00}  & 87.20  & 76.10  & 62.25  & 96.00  & \textbf{95.20}  & \textbf{93.90} & \textbf{91.25}  \\
      \midrule
        \multirow{3}{*}{TR} & R@1 & 18.00  & 7.60  & 4.60  & 3.15  & 70.00  & 40.40  & 26.60  & 18.45  & \textbf{90.00}  & \textbf{78.40}  & \textbf{71.30}  & \textbf{63.90}  \\ 
        & R@5 & 47.00  & 20.20  & 13.00  & 8.55  & \textbf{95.00}  & 74.20  & 61.80  & 47.30  & \textbf{95.00}  & \textbf{93.40}  & \textbf{91.40}  & \textbf{88.05}  \\ 
        & R@10 & 59.00  & 31.00  & 19.80  & 13.55  & \textbf{98.00}  & 87.00  & 75.30  & 62.15  & 95.00  & \textbf{95.60}  & \textbf{94.10}  & \textbf{91.90} \\
      \bottomrule
    \end{tabular}
    \label{tab:retrieval_main}
\end{table*}

\subsection{Radiology Report Generation}

\subsubsection{Quantitative Performance Evaluation}
Med3DVLM achieves state-of-the-art performance in radiology report generation, as shown in Table~\ref{tab:report_eval}. It outperforms all baselines across BLEU, ROUGE, and METEOR, with particularly notable gains in METEOR, which captures both fluency and content relevance. These results suggest that Med3DVLM can generate clinically coherent, well-structured descriptions of complex imaging findings.


Interestingly, although Med3DVLM and M3D-LaMed achieve similar BERTScores (88.11 vs. 88.46), BERTScore primarily reflects abstract similarity in latent space and does not fully capture linguistic precision or clinical completeness. In contrast, Med3DVLM shows substantial gains on semantically grounded metrics, with METEOR improving from 14.38\% to 36.42\% and ROUGE from 19.55\% to 40.25\%. These improvements indicate that Med3DVLM generates reports with more accurate lexical choices, clearer structure, and better alignment with clinical content. The discrepancy between BERTScore and traditional n-gram metrics highlights the importance of evaluating both semantic fidelity and textual quality, which are essential for accurate and readable medical reporting.

\begin{table}[!ht]
  \caption{Report generation performance on the M3D-Cap dataset.}
  \centering
  \footnotesize
  \begin{tabular}{l|cccc}
    \toprule
      Method & BLEU & ROUGE & METEOR & BERTScore \\
    \midrule
      RadFM & 12.23 & 16.49 & 11.57 & 87.93 \\
      M3D-LaMed (Linear) & 14.49 & 19.25 & 14.11 & 88.32 \\
      M3D-LaMed (MLP) & 15.15 & 19.55 & 14.38 & \textbf{88.46} \\
    \midrule
      \makecell[l]{Med3DVLM (Ours)} & \textbf{36.88} & \textbf{40.25} & \textbf{36.42} & 88.11 \\
    \bottomrule
  \end{tabular}
  \label{tab:report_eval}
\end{table}

\subsubsection{Qualitative Analysis of Generated Reports}
\begin{figure*}[!ht]
  \centering
  \includegraphics[width=0.9\textwidth]{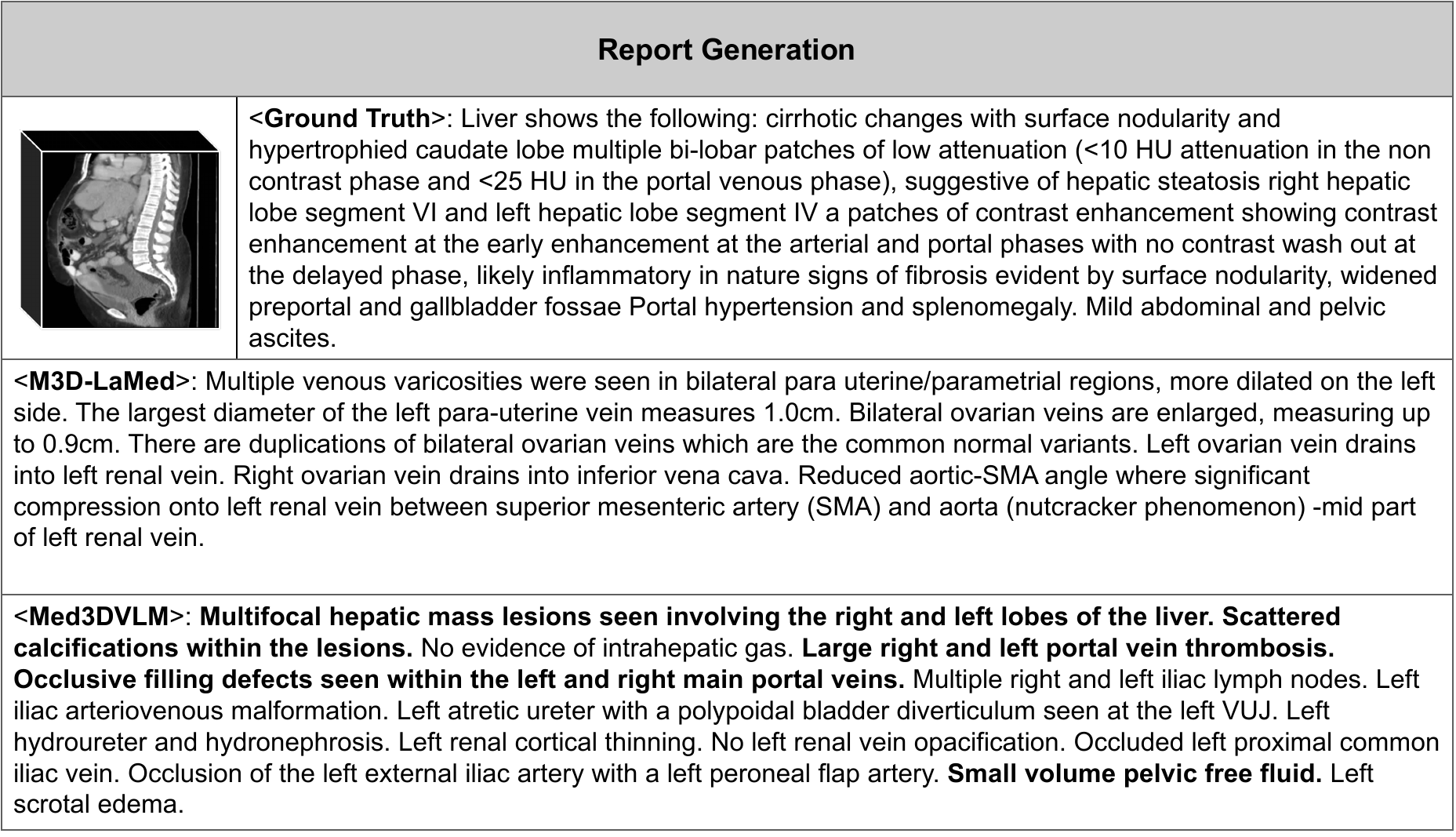}
  \caption{An example of generated radiology reports by Med3DVLM and M3D-LaMed. Med3DVLM demonstrates improved alignment with clinical findings, while M3D-LaMed generates irrelevant or incorrect content.}
  \label{fig:report_examples}
\end{figure*}

Figure~\ref{fig:report_examples} compares the radiology reports generated by Med3DVLM and M3D-LaMed on a chest CT. Med3DVLM correctly identifies key abnormalities, including multifocal hepatic mass lesions, portal vein thrombosis, and occlusive filling defects, aligning well with the ground truth. It also captures scattered calcifications and small-volume pelvic free fluid, findings not explicitly mentioned but plausibly related to the documented pathology. In contrast, M3D-LaMed fails to detect any liver pathology, underscoring its limitations in aligning imaging features with diagnostic text.

Despite its improvements, Med3DVLM generates incorrect findings, such as hepatic steatosis, contrast enhancement, and iliac arteriovenous malformation, suggesting a tendency to overgeneralize common radiological patterns. Mentions of left ureter abnormalities and renal cortical thinning appear to be hallucinated, indicating the need for better factual grounding. M3D-LaMed, while not generating outright hallucinations, produces findings entirely unrelated to liver pathology, such as bilateral ovarian vein enlargement and para-uterine venous varicosities, suggesting weaker cross-modal alignment.

\subsection{Visual Question Answering}

\subsubsection{Quantitative Performance Evaluation}


Med3DVLM achieves state-of-the-art performance in both open-ended and close-ended VQA tasks, as shown in Tables~\ref{tab:open_vqa_eval} and~\ref{tab:close_vqa_eval}. The model demonstrates particularly strong results in clinically important categories such as organ and abnormality identification, highlighting its ability to capture fine-grained anatomical and contextual cues in 3D volumes.

In open-ended VQA, Med3DVLM improves the METEOR score from 33.58\% to 36.76\% (a 9.5\% improvement) and the ROUGE score from 52.39\% to 56.31\% (a 7.5\% improvement) compared to M3D-LaMed. These metrics offer a more semantically grounded evaluation than simple lexical overlap, capturing both the accuracy and fluency of generated responses, which are essential for answering complex clinical queries with clarity and relevance. The results suggest that Med3DVLM produces free-text responses that are not only better aligned with the ground truth but also more structured and clinically coherent.

In close-ended VQA, Med3DVLM achieves 79.75\% accuracy, surpassing M3D-LaMed’s 75.78\% by 5.2\%. Although the margin is smaller than in open-ended VQA, the improvement remains clinically meaningful: even modest gains in accuracy can reduce diagnostic errors and enhance the reliability of automated decision support systems in clinical practice.

\begin{table*}[!ht]
  \caption{Open-ended visual question answering performance on the M3D-VQA dataset.}
  \centering
  \small
  \begin{tabular}{c|l|ccccc|c}
  \hline
    \toprule
      \multicolumn{8}{c}{Open-ended VQA} \\
    \midrule
      Method & Metric & Plane & Phase & Organ & Abnormality & Location & Mean \\
    \midrule
      \multirow{4}{*}{RadFM} & BLEU & 14.24 & 14.25 & 14.24 & 15.64 & 23.58 & 16.39 \\
      & ROUGE & 25.40 & 25.41 & 25.38 & 25.38 & 29.09 & 26.13 \\
      & METEOR & 20.62 & 20.63 & 20.61 & 20.60 & 24.19 & 21.33 \\
      & BERTScore & 92.68 & 92.04 & 86.79 & 85.84 & 86.26 & 88.72 \\
    \midrule
      \multirow{4}{*}{M3D-LaMed} & BLEU & 98.37 & 74.41 & 34.20 & 15.91 & 24.00 & 49.38 \\
      & ROUGE & 98.42 & 78.63 & 37.87 & 19.27 & 27.74 & 52.39 \\
      & METEOR & 49.20 & 63.58 & 23.78 & 12.83 & 18.50 & 33.58 \\
      & BERTScore & 99.47 & 95.55 & 88.97 & 86.08 & 87.60 & 91.53 \\
    \midrule
      \multirow{4}{*}{Med3DVLM (Ours)} & BLEU & \bfseries 98.85  & \bfseries 78.17  & \bfseries 40.22  & \bfseries 18.99  & \bfseries 25.66  & \bfseries 52.38  \\
      & ROUGE & \bfseries 98.89  & \bfseries 84.20  & \bfseries 45.22  & \bfseries 23.27  & \bfseries 29.99  & \bfseries 56.31  \\
      & METEOR & \bfseries 49.43  & \bfseries 68.50  & \bfseries 29.32  & \bfseries 16.21  & \bfseries 20.32  & \bfseries 36.76  \\
      & BERTScore & \bfseries 99.83  & \bfseries 96.47  & \bfseries 90.47  & \bfseries 86.27  & \bfseries 87.88  & \bfseries 92.18 \\
    \bottomrule
  \end{tabular}
  \label{tab:open_vqa_eval}
\end{table*}

\begin{table*}[!ht]
  \caption{Comparison of closed-ended visual question answering performance with state-of-the-art methods on the M3D-VQA dataset. The results are reported in terms of accuracy.}
  \centering
  \small
  \begin{tabular}{l|ccccc|c}
  \hline
    \toprule
      \multicolumn{7}{c}{Close-ended VQA} \\
    \midrule
      Methods & Plane & Phase & Organ & Abnormality & Location & Mean \\
    \midrule
      RadFM & 19.65  & 28.70  & 16.80  & 18.92  & 14.88  & 19.79  \\
      M3D-LaMed & 98.80  & 79.75  & 74.75  & 66.65  & 58.94  & 75.78  \\ 
    \midrule
      Med3DVLM (Ours) & \bfseries 99.15 & \bfseries 87.50 & \bfseries 77.45 & \bfseries 70.17 & \bfseries 64.49 & \bfseries 79.75 \\
    \bottomrule
  \end{tabular}
  \label{tab:close_vqa_eval}
\end{table*}

\subsubsection{Qualitative Analysis of VQA Accuracy}
\begin{figure*}[!ht]
  \centering
  \includegraphics[width=0.9\textwidth]{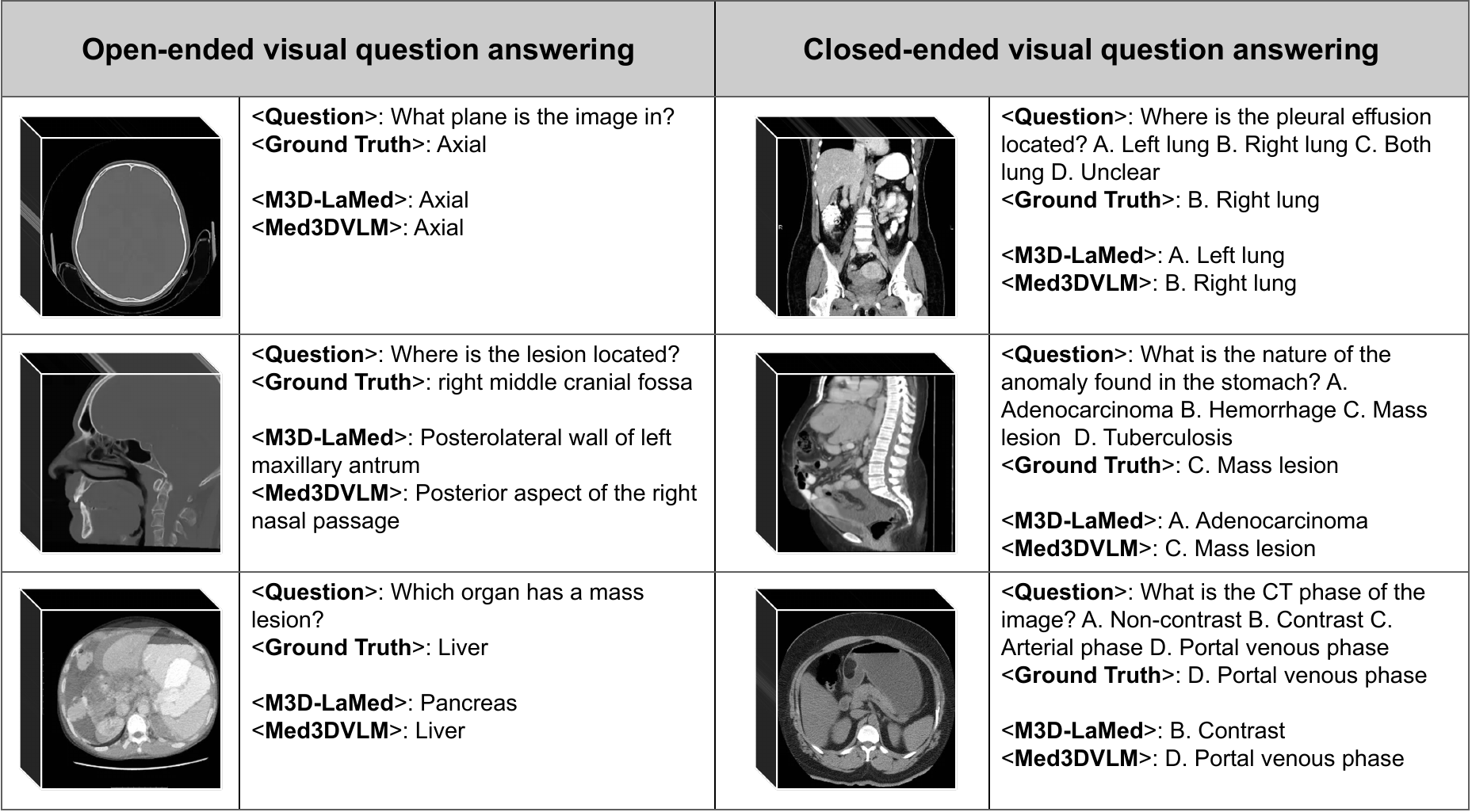}
  \caption{Examples of open-ended and closed-ended VQA results by Med3DVLM and M3D-LaMed.}
  \label{fig:vqa_examples}
\end{figure*}

Figure~\ref{fig:vqa_examples} presents a qualitative comparison of Med3DVLM and M3D-LaMed in open-ended and closed-ended VQA. In open-ended VQA, both M3D-LaMed and Med3DVLM correctly recognizes the axial imaging plane and  localizes a mass lesion in the liver. However, in more complex cases, such as lesion localization in the cranial fossa, Med3DVLM provides a partially correct answer, whereas M3D-LaMed misidentifies the region entirely.  While Med3DVLM significantly improves answer accuracy, errors in lesion localization indicate room for enhancement in fine-grained spatial reasoning and uncertainty calibration.

In closed-ended VQA, Med3DVLM consistently outperforms M3D-LaMed by correctly identifying the pleural effusion in the right lung, the mass lesion in the stomach, and the portal venous phase of the CT scan. In contrast, M3D-LaMed misclassifies the pleural effusion as being in the left lung, the stomach lesion as adenocarcinoma, and the CT phase as contrast phase. These errors indicate M3D-LaMed's limitations in recognizing imaging features and aligning them with clinical knowledge.

\subsection{Ablation Study}

\subsubsection{Effect of Core Components}
\begin{table*}[!ht]
\caption{Effect of core Med3DVLM components on radiology report generation.}
\centering
\small
\begin{tabular}{ccc|cccc}
\toprule
DCFormer            & SigLIP Loss         & MLP-Mixer           & \multicolumn{1}{c}{BLEU} & \multicolumn{1}{c}{ROUGE} & \multicolumn{1}{c}{METEOR} & \multicolumn{1}{c}{BERT} \\ \midrule
\checkmark          &                     & \checkmark          & 24.47                    & 28.80                     & 23.78                      & 85.75                    \\
\checkmark          & \checkmark          &                     & 14.31                    & 18.14                     & 13.48                      & 83.86                    \\
                    & \checkmark          &                     & 8.62                     & 11.23                     & 8.88                       & 77.07                    \\
\textbf{\checkmark} & \textbf{\checkmark} & \textbf{\checkmark} & \textbf{36.88}                    & \textbf{40.25}                     & \textbf{36.42}                      & \textbf{88.11}                    \\ \bottomrule
\end{tabular}
\label{tab:full_report_ablation}
\end{table*}

\begin{table*}[]
\caption{Effect of core Med3DVLM components on open-ended VQA performance.}
\centering
\small
\begin{tabular}{ccc|l|ccccc|l}
\toprule
DCFormer                    & SigLIP Loss                 & MLP-Mixer                   & Metric & \multicolumn{1}{c}{Plane} & \multicolumn{1}{c}{Phase} & \multicolumn{1}{c}{Organ} & \multicolumn{1}{c}{Abnormality} & \multicolumn{1}{c}{Location} & \multicolumn{1}{c}{Mean} \\ \midrule
\multirow{4}{*}{\checkmark} & \multirow{4}{*}{}           & \multirow{4}{*}{\checkmark} & BLEU   & \textbf{99.00}                     & 77.28                     & 39.13                     & \textbf{19.73}                           & 24.85                        & 52.00                    \\
                            &                             &                             & ROUGE  & \textbf{99.05}                     & \textbf{84.23}                     & 43.48                     & \textbf{23.85}                           & 29.48                        & 56.02                    \\
                            &                             &                             & METEOR & \textbf{49.52}                     & 68.31                     & 27.23                     & \textbf{16.69}                           & 20.07                        & 36.36                    \\
                            &                             &                             & BERT   & \textbf{99.85}                     & 96.30                     & 90.33                     & \textbf{86.29}                           & 87.74                        & 92.10                    \\ \midrule
\multirow{4}{*}{\checkmark} & \multirow{4}{*}{\checkmark} & \multirow{4}{*}{}           & BLEU   & 98.60                     & 75.00                     & 34.34                     & 17.76                           & 24.55                        & 50.05                    \\
                            &                             &                             & ROUGE  & 98.67                     & 81.55                     & 38.59                     & 21.45                           & 28.84                        & 53.82                    \\
                            &                             &                             & METEOR & 49.37                     & 65.18                     & 24.28                     & 14.72                           & 19.69                        & 34.65                    \\
                            &                             &                             & BERT   & 99.79                     & 95.87                     & 89.44                     & 85.93                           & 87.71                        & 91.75                    \\ \midrule
\multirow{4}{*}{}           & \multirow{4}{*}{\checkmark} & \multirow{4}{*}{}           & BLEU   & 0.00                      & 0.17                      & 0.19                      & 0.73                            & 0.57                         & 0.33                     \\
                            &                             &                             & ROUGE  & 0.08                      & 0.43                      & 0.96                      & 1.54                            & 1.46                         & 0.89                     \\
                            &                             &                             & METEOR & 0.07                      & 0.90                      & 1.62                      & 2.19                            & 2.14                         & 1.38                     \\
                            &                             &                             & BERT   & 73.18                     & 74.68                     & 74.47                     & 74.01                           & 74.24                        & 74.12                    \\ \midrule
\multirow{4}{*}{\checkmark} & \multirow{4}{*}{\checkmark} & \multirow{4}{*}{\checkmark} & BLEU   & 98.85                     & \textbf{78.17}                     & \textbf{40.22}                     & 18.99                           & \textbf{25.66}                        & \textbf{52.38}                    \\
                            &                             &                             & ROUGE  & 98.89                     & 84.20                     & \textbf{45.22}                     & 23.27                           & \textbf{29.99}                        & \textbf{56.31}                    \\
                            &                             &                             & METEOR & 49.43                     & \textbf{68.50}                     & \textbf{29.32}                     & 16.21                           & \textbf{20.32}                        & \textbf{36.76}                    \\
                            &                             &                             & BERT   & 99.83                     & \textbf{96.47}                     & \textbf{90.47}                     & 86.27                           & \textbf{87.88}                        & \textbf{92.18}                   \\ \bottomrule
\end{tabular}
\label{tab:all_open_vqa_ablation}
\end{table*}

\begin{table*}[!ht]
\caption{Effect of core Med3DVLM components on closed-ended VQA accuracy.}
\centering
\small
\begin{tabular}{ccc|ccccc|c}
\toprule
DCFormer   & SigLIP     & MLP-Mixer  & \multicolumn{1}{c}{Plane} & \multicolumn{1}{c}{Phase} & \multicolumn{1}{c}{Organ} & \multicolumn{1}{c}{Abnormality} & \multicolumn{1}{c}{Location} & \multicolumn{1}{c}{Mean} \\ \midrule
\checkmark &            & \checkmark & \textbf{99.30}                     & \textbf{87.95}                     & 77.40                     & 69.90                           & 63.99                        & 79.71                    \\
\checkmark & \checkmark &            & 98.85                     & 84.95                     & 74.80                     & 69.82                           & 64.46                        & 78.58                    \\
           & \checkmark &            & 1.60                      & 1.05                      & 1.35                      & 1.03                            & 1.28                         & 1.26                     \\
\checkmark & \checkmark & \checkmark & 99.15                     & 87.50                     & \textbf{77.45}                     & \textbf{70.17}                           & \textbf{64.49}                        & \textbf{79.75}                   \\ \bottomrule
\end{tabular}
\label{tab:full_close_vqa_ablation}
\end{table*}

We first analyze the impact of the three core components of Med3DVLM: the DCFormer vision encoder, the SigLIP contrastive loss, and the multi-scale MLP-Mixer projector. The results are summarized in Table~\ref{tab:full_report_ablation}, Table~\ref{tab:all_open_vqa_ablation}, and Table~\ref{tab:full_close_vqa_ablation}, corresponding to radiology report generation, open-ended VQA, and closed-ended VQA, respectively.

In the radiology report generation task, the full model achieves a METEOR score of 36.42\%. Removing the MLP-Mixer leads to a drop in performance to 23.78\%, while removing the SigLIP loss results in a lower METEOR score of 13.48\%. When the DCFormer encoder is removed, the MLP-Mixer must also be excluded since it depends on the multi-scale feature outputs from DCFormer. In this case, the METEOR score degrades severely, reaching only 8.88\%. Similar trends are observed in BLEU, ROUGE, and BERTScore, highlighting the critical role of all three components in producing clinically accurate and semantically aligned text.

For open-ended VQA, removing both DCFormer and MLP-Mixer leads to a drastic drop in performance, reducing the mean METEOR score from 36.76\% to just 1.38\%. In contrast, removing only the MLP-Mixer causes a modest drop to 34.65\%, while removing SigLIP results in a minimal decrease to 36.36\%. Similarly, for closed-ended VQA, the full model achieves a mean accuracy of 79.75\%. Removing SigLIP has little effect (79.71\%), and removing the MLP-Mixer causes a small drop to 78.58\%. In contrast, removing both DCFormer and MLP-Mixer results in a sharp decline to 1.26\% accuracy. These results suggest that while all three components contribute to overall VQA performance, the DCFormer encoder is essential for reliable visual understanding, and the MLP-Mixer plays an important role in multi-modal reasoning.

In summary, each core component of Med3DVLM contributes to overall performance, with varying importance across tasks. The DCFormer encoder is essential for robust 3D representation across all settings, while the MLP-Mixer plays a key role in multi-modal reasoning, particularly for generative tasks. The SigLIP loss enhances alignment stability, especially in report generation.

\subsubsection{Impact of Vision Encoder and Contrastive Loss}
\begin{table*}[!ht]
  \caption{The impact of different vision encoders and loss functions on image-text retrieval. DCFormer-S represents small version of DCFormer.}
  \centering
  \small
  \begin{tabular}{c|c|rrrr|rrrr}
  \hline
    \toprule
      \multicolumn{2}{c|}{Methods} & \multicolumn{4}{c|}{ViT3D CLIP} & \multicolumn{4}{c}{ViT3D SigLIP} \\
    \midrule
      \multicolumn{2}{c|}{Test Samples} & 100 & 500 & 1000 & 2000 & 100 & 500 & 1000 & 2000\\
    \midrule
      \multirow{3}{*}{IR} & R@1 & 63.00  & 33.40  & 25.90  & 19.10 & 75.00 & 55.40 & 46.20 & 38.65 \\
      & R@5 & 89.00  & 67.80  & 57.20  & 44.70 & 91.00 & 84.40 & 79.50 & 70.50 \\
      & R@10 & 95.00  & 81.00  & 71.60  & 59.35 & 94.00 & 89.80 & 87.10& 79.45 \\
    \midrule
      \multirow{3}{*}{TR} & R@1 & 57.00  & 32.00  & 25.50  & 19.40 & 78.00 & 59.60 & 49.40 & 39.70 \\
      & R@5 & 90.00  & 67.40  & 56.90  & 45.35 & 91.00 & 84.20 & 77.40 & 69.20 \\
      & R@10 & 94.00  & 81.80  & 70.70  & 59.10 & 93.00 & 88.80 & 85.50 & 78.40 \\
    \midrule
    \midrule
      \multicolumn{2}{c|}{Methods} & \multicolumn{4}{c|}{DCFormer-S CLIP} & \multicolumn{4}{c}{DCFormer-S SigLIP} \\
    \midrule
      \multicolumn{2}{c|}{Test Samples} & 100 & 500 & 1000 & 2000 & 100 & 500 & 1000 & 2000\\
    \midrule
      \multirow{3}{*}{IR} & R@1 & 76.00  & 51.80  & 40.80  & 29.85  & \bfseries 92.00  & \bfseries 77.20  & \bfseries 69.30  & \bfseries 61.00  \\
      & R@5 & 94.00  & 86.00  & 74.60  & 64.60  & \bfseries 95.00  & \bfseries 92.80  & \bfseries 91.40  & \bfseries 87.20  \\
      & R@10 & \bfseries 97.00  & 91.40  & 84.90  & 75.95  & 96.00  & \bfseries 95.20  & \bfseries 93.90  & \bfseries 91.25  \\
    \midrule
      \multirow{3}{*}{TR} & R@1 & 83.00  & 52.60  & 43.50  & 32.60  & \bfseries 90.00  & \bfseries 78.40  & \bfseries 71.30  & \bfseries 63.90  \\
      & R@5 & 94.00  & 86.00  & 75.00  & 64.10  & \bfseries 95.00  & \bfseries 93.40  & \bfseries 91.40  & \bfseries 88.05  \\
      & R@10 & \bfseries 98.00  & 91.60  & 85.70  & 76.70  & 95.00  & \bfseries 95.60  & \bfseries 94.10  & \bfseries 91.90 \\
    \bottomrule
  \end{tabular}
  \label{tab:retrieval_ablation}
\end{table*}

We evaluated the impact of different vision encoders (ViT3D vs. DCFormer-S) and contrastive loss functions (CLIP vs. SigLIP) on image-text retrieval. As shown in Table~\ref{tab:retrieval_ablation}, both image retrieval (IR) and text retrieval (TR) benefit significantly from using DCFormer-S over ViT3D, particularly at larger test sizes. For example, with 2,000 test samples and SigLIP loss, DCFormer-S achieves an R@1 of 61.00\% for IR and 63.90\% for TR, compared to 38.65\% and 39.70\% respectively with ViT3D.

Across all encoder types and test sizes, switching from CLIP to SigLIP consistently improves performance. For example, with the ViT3D encoder and 2,000 test samples, SigLIP improves IR R@1 from 19.10\% to 38.65\%, and TR R@1 from 19.40\% to 39.70\%. With the DCFormer-S encoder, the gains are also evident: IR R@1 increases from 29.85\% to 61.00\%, and TR R@1 from 32.60\% to 63.90\%. These improvements are especially pronounced at larger evaluation scales, where SigLIP enhances alignment stability without requiring large negative batches.

\begin{table*}[!ht]
  \caption{The impact of multi-modal projectors on radiology report generation. All models use DCFormer-small and the Qwen-2.5-7B-Instruct LLM. H: low high hybrid.}
  \centering
 \small
  \begin{tabular}{l|cccc|cc}
    \toprule
      Method & BLEU & ROUGE & METEOR & BERTScore & Parameters & Flops \\
    \midrule
      2xMLP & 15.63 & 19.86 & 15.10 & 84.24 & 15.60 M & 0.50 G \\
      2xMLP-H & 15.99 & 20.25 & 15.68 & 84.31 & 16.98 M & 4.14 G \\
      1xMLP-Mixer-H & 23.93 & 27.70 & 23.25 & 85.73 & 29.91 M & 3.85 G \\
    \midrule
      2xMLP-Mixer-H & \bfseries 36.88 & \bfseries 40.25 & \bfseries 36.42 & \bfseries 88.11 & 47.22 M & 6.14 G\\
    \bottomrule
  \end{tabular}
  \label{tab:multi_modal_proj_report}
\end{table*}

\subsubsection{Impact of Multi-Modal Projectors}
\begin{table*}[!ht]
  \caption{The impact of multi-modal projectors on open-ended visual question answering. All models used the DCFormer-small vision encoder and Qwen 2.5-7B-Instruct model. H: low high hybrid.}
  \centering
  \small
  \begin{tabular}{c|l|ccccc|c}
  \hline
    \toprule
      \multicolumn{8}{c}{Open-ended VQA} \\
    \midrule
      Method & Metric & Plane & Phase & Organ & Abnormality & Location & Mean \\
    \midrule
      \multirow{4}{*}{2xMLP} & BLEU & 98.67 & 74.23 & 33.95 & 16.92 & 23.79 & 49.51 \\
      & ROUGE & 98.72 & 80.72 & 38.46 & 20.97 & 28.04 & 53.38 \\
      & METEOR & 49.35 & 64.69 & 24.30 & 14.63 & 18.97 & 34.39 \\
      & BERTScore & 99.80 & 95.75 & 89.50 & 85.88 & 87.64 & 91.71 \\
    \midrule
      \multirow{4}{*}{2xMLP-H} & BLEU & \bfseries 98.86 & 74.46 & 34.37 & 16.96 & 24.49 & 49.83 \\
      & ROUGE & 98.91 & 80.87 & 38.55 & 20.82 & 28.81 & 53.59 \\
      & METEOR & 49.48 & 64.88 & 24.42 & 14.36 & 19.29 & 34.49 \\
      & BERTScore & \bfseries 99.83 & 95.75 & 89.50 & 85.88 & 87.80 & 91.75 \\
    \midrule
      \multirow{4}{*}{1xMLP-Mixer-H} & BLEU & 98.85  & 76.37  & 38.87  & 17.47  & 24.62  & 51.24  \\
      & ROUGE & \bfseries 98.93  & 82.69  & 42.93  & 21.46  & 28.94  & 54.99  \\
      & METEOR & \bfseries 49.51  & 67.00  & 26.86  & 15.19  & 19.63  & 35.64 \\
      & BERTScore & \bfseries 99.83  & 96.16  & 90.29  & 85.99  & 87.80  & 92.01 \\
    \midrule
      \multirow{4}{*}{2xMLP-Mixer-H} & BLEU & 98.85  & \bfseries 78.17  & \bfseries 40.22  & \bfseries 18.99  & \bfseries 25.66  & \bfseries 52.38  \\
      & ROUGE & 98.89  & \bfseries 84.20  & \bfseries 45.22  & \bfseries 23.27  & \bfseries 29.99  & \bfseries 56.31  \\
      & METEOR &  49.43  & \bfseries 68.50  & \bfseries 29.32  & \bfseries 16.21  & \bfseries 20.32  & \bfseries 36.76  \\
      & BERTScore & \bfseries 99.83  & \bfseries 96.47  & \bfseries 90.47  & \bfseries 86.27  & \bfseries 87.88  & \bfseries 92.18 \\
    \bottomrule
  \end{tabular}
  \label{tab:multi_modal_proj_open}
\end{table*}

\begin{table*}[!ht]
  \caption{The impact of multi-modal projectors on closed-ended visual question answering. All models used the DCFormer-small vision encoder and Qwen 2.5-7B-Instruct LLM. H: low high hybrid.}
  \centering
  \small
  \begin{tabular}{l|ccccc|c}
  \hline
    \toprule
      \multicolumn{7}{c}{Close-ended VQA} \\
    \midrule
      Methods & Plane & Phase & Organ & Abnormality & Location & Mean \\
    \midrule
      2xMLP & 98.75  & 84.35  & 75.90  & 69.55  & 64.70  & 78.65  \\
      2xMLP-H & 98.75  & 86.40  & 75.05  & 69.90  & 62.67  & 78.55  \\ 
      1xMLP-Mixer-H & \bfseries 99.30 & 86.10 & 77.05 & \bfseries 70.19 & 63.92 & 79.31 \\
    \midrule
      2xMLP-Mixer-H & 99.15 & \bfseries 87.50 & \bfseries 77.45 & 70.17 & \bfseries 64.49 & \bfseries 79.75 \\
    \bottomrule
  \end{tabular}
  \label{tab:multi_modal_proj_close}
\end{table*}

We evaluated the impact of different multi-modal projector designs on radiology report generation, open-ended VQA, and closed-ended VQA. 
The 2×MLP variant consists of a simple two-layer multilayer perceptron applied to global visual features, using a linear activation followed by GELU and another linear transformation. The 2×MLP-H variant incorporates hierarchical features from the final and penultimate layers of the vision encoder; each feature map is independently processed by a 2×MLP block, and their outputs are concatenated. The 1×MLP-Mixer-H design replaces the MLPs with a single MLP-Mixer block that jointly processes the concatenated hierarchical features, enabling token and channel mixing. The 2×MLP-Mixer-H configuration, used in our final model, stacks two MLP-Mixer blocks and includes an additional hidden projection layer to further enhance multi-scale feature interaction. To provide a complete view of computational efficiency, we also report the number of parameters and FLOPs for each projector variant in Table~\ref{tab:multi_modal_proj_report}

As shown in Table~\ref{tab:multi_modal_proj_report}, the 2$\times$MLP-Mixer-H configuration achieves the highest METEOR score of 36.42\% on report generation, significantly outperforming simpler alternatives such as 2$\times$MLP (15.10\%). The single-scale 1$\times$MLP-Mixer-H achieves a much higher METEOR of 23.25\% compared to MLP-based projectors, although it still lags behind the proposed multi-scale design. These results highlight the importance of multi-scale feature integration for generating coherent and semantically aligned clinical reports.

For both open-ended and closed-ended VQA, the 2$\times$MLP-Mixer-H projector achieves the best overall performance (Tables~\ref{tab:multi_modal_proj_open} and~\ref{tab:multi_modal_proj_close}). In open-ended VQA, it improves the mean METEOR from 34.39\% (2$\times$MLP) to 36.76\%, with strong gains in semantically challenging categories like “organ” and “abnormality.” In closed-ended VQA, although the performance gap is narrower, the 2$\times$MLP-Mixer-H still achieves the highest mean accuracy of 79.75\%, compared to 78.65\% with 2$\times$MLP. These results suggest that while classification tasks rely more on pattern recognition, generative tasks benefit significantly from multi-scale feature fusion, and the MLP-Mixer projector enhances semantic reasoning and alignment across both task types.

\section{Discussion}

\subsection{Addressing Limitations of Existing Models}
While previous 3D VLMs such as M3D-LaMed and RadFM have made significant advances, they remain constrained by computational inefficiency, limited capacity for fine-grained spatial reasoning, and suboptimal multi-modal alignment. Med3DVLM addresses these gaps through three targeted innovations. First, Med3DVLM uses DCFormer to decompose 3D convolutions to reduce complexity while preserving spatial detail, allowing for the processing of high-resolution image volumes. Second, SigLIP improves contrastive learning on small, semantically dense medical datasets by avoiding reliance on large negative batches. Third, the dual-stream MLP-Mixer projector effectively fuses low- and high-level image features with text embeddings, enriching the semantic alignment with the LLM.

\subsection{Clinical Implications}
The integration of Med3DVLM into clinical workflows offers strong potential to improve medical imaging interpretation, decision support, and patient care. Its robust performance in image-text retrieval enables accurate alignment between 3D imaging studies and corresponding textual reports, facilitating rapid access to similar prior cases. This capability supports comparative analysis, longitudinal tracking, and second-opinion workflows, and can enhance diagnostic accuracy and consistency. In educational settings, it may also serve as a valuable tool for case-based learning and clinical training. Med3DVLM also generates accurate, fluent radiology reports directly from 3D volumetric data, reducing documentation burden—especially in high-volume clinical environments. Its substantial gains in METEOR and ROUGE scores indicate strong alignment with expert-level interpretations, making it a practical tool for assisting in preliminary report drafting and standardizing report quality. Finally, the model’s VQA functionality allows clinicians to interact with 3D scans using natural language. This enables task-specific, context-aware responses that support anatomical identification, diagnostic clarification, and real-time clinical decision-making—ultimately reducing the risk of oversight and enhancing confidence in image interpretation.

\subsection{Challenges and Future Directions}
While Med3DVLM sets a new standard in 3D VLM, several challenges remain. The model occasionally generates hallucinated content in radiology reports, such as references to unrelated anatomical findings, highlighting the persistent issue of factual grounding in generative VLMs. Additionally, although performance in open-ended VQA improved markedly, some errors in lesion localization and answer precision persist, suggesting the need for enhanced spatial reasoning and more robust alignment between visual features and textual outputs.

While the M3D dataset provides a large-scale and diverse benchmark, it primarily consists of CT images and focuses on a limited set of tasks including retrieval, report generation, and VQA. To ensure broader generalization and real-world applicability, future evaluation should incorporate other imaging modalities such as MRI and ultrasound, as well as datasets that differ in patient demographics, clinical settings, and scanner protocols. In addition, extending the scope of evaluation to include tasks such as medical image synthesis, modality translation, and segmentation would offer a more comprehensive assessment of the model's robustness and clinical utility across a wider range of scenarios.

As for future directions, incorporating structured clinical knowledge could help reduce hallucinations and improve interpretability. Furthermore, as noted above, validating Med3DVLM across diverse clinical datasets and imaging protocols will be critical to ensuring its reliability in real-world settings. Expanding the model to support uncertainty-aware response generation may also enhance its trustworthiness in clinical decision support.

\subsection{Computational Complexity}

We compare Med3DVLM and M3D-LaMed in terms of parameter count and FLOPs, as summarized in Table~\ref{tab:total_params}. Med3DVLM replaces the 3D Vision Transformer in M3D-LaMed with the DCFormer encoder, reducing the vision backbone from 87.4M parameters and 253.23G FLOPs to 18.2M parameters and 21.59G FLOPs. Although Med3DVLM uses a larger multi-modal projector (47.2M vs. 19.9M parameters), this overhead is minor compared to the efficiency gains from the encoder. The total model size increases slightly (7.6B vs. 6.9B) due to a larger LLM, which is fine-tuned efficiently using LoRA.

We do not report direct comparisons of inference time and memory usage, as M3D-LaMed includes an additional segmentation module not present in our model. This makes runtime comparisons non-equivalent. To ensure fairness, we focus on theoretical complexity metrics that isolate the vision-language components.

\begin{table}[!ht]
\small
\centering
\caption{Comparison of parameters and FLOPs for each module in Med3DVLM. M3D-LaMed values are reported from the original paper.}
\begin{tabular}{l|llll}
\toprule
\multirow{2}{*}{Module} & \multicolumn{2}{c}{M3D-LaMed}  & \multicolumn{2}{c}{Med3DVLM} \\ \cmidrule{2-5}
                     & \multicolumn{1}{c}{Params} & \multicolumn{1}{c}{FLOPs} & \multicolumn{1}{c}{Params} & \multicolumn{1}{c}{FLOPs} \\
                     \midrule
3D Image Encoder     & 87.4M & 253.23G      & 18.2M & 21.59G   \\
Multimodal Projector & 19.9M & 5.10G    & 47.2M & 6.14G   \\
LLM with LoRA        & 6.7B  & -     & 7.6B & -    \\ \midrule
All                  & 6.9B  & -     & 7.6B & -   \\ \bottomrule
\end{tabular}
\label{tab:total_params}
\end{table}

\section{Conclusion}

We presented Med3DVLM, a vision-language model designed specifically for 3D medical image analysis. Med3DVLM integrates an efficient volumetric encoder (DCFormer), a sigmoid-based contrastive learning strategy (SigLIP), and a dual-stream MLP-Mixer projector. Extensive experiments on the M3D dataset demonstrate that Med3DVLM achieves state-of-the-art performance in image-text retrieval, radiology report generation, and visual question answering. Notably, Med3DVLM maintains high accuracy while remaining computationally efficient, making it suitable for deployment in real-world clinical workflows. These results underscore the promise of 3D VLMs as foundational tools for building automated, interpretable, and multimodal medical AI systems.

\FloatBarrier

{\small
\bibliographystyle{ieee_fullname}
\bibliography{ref}
}

\end{document}